\documentclass{article}

\usepackage{spconf_new}

\usepackage[hyphens]{url}
\usepackage{hyperref}
\usepackage[hyphenbreaks]{breakurl}
\usepackage{amsmath,graphicx}
\usepackage{amssymb,bm}
\usepackage{amsfonts}
\usepackage{tabularx, booktabs}
\usepackage{bbm}
\usepackage{subcaption}  
\usepackage[justification=centering]{caption}
\newcolumntype{Y}{>{\centering\arraybackslash}X}
\usepackage{multirow}

\graphicspath{{../figures/}}

\usepackage{tikz,placeins}
\usetikzlibrary{arrows,decorations.markings}
\tikzstyle{every picture}+=[font=\rmfamily\it\bfseries\large]
\usetikzlibrary{positioning, shapes}
\pgfdeclarelayer{background}
\pgfdeclarelayer{foreground}
\pgfsetlayers{background,main,foreground}

\usepackage{algorithm}
\usepackage{algpseudocode}

\newcommand{\specialcell}[2][c]{%
  \begin{tabular}[#1]{@{}c@{}}#2\end{tabular}}

\newcommand{\ssf}[1]{%
  \bm{\mathsf{#1}}%
}

\title{Macro-block dropout for improved regularization in training end-to-end speech recognition models}

\name{Chanwoo Kim$^{1}$, Sathish Indurti$^{1}$, Jinhwan Park$^{1}$, and Wonyong Sung$^{2}$
 \thanks{Thanks to Samsung
Electronics for funding this research. The authors are thankful to
president Sebastian Seung, Executive Vice President (EVP) Daniel Lee, EVP Seunghwan
Cho, and the Language and Voice Team (LVT) members at Samsung Research.}
}
\address{Samsung Research$^{1}$, Seoul, South Korea \\
         Seoul National University$^{2}$, Seoul, South Korea \\
      {\small \tt \{chanw.com, s.indurti, jh0354.park\}@samsung.com, wysung@snu.ac.kr}}

\begin{document}

\maketitle
\begin{abstract}
  This paper proposes a new regularization algorithm referred to as {\it macro-block
  dropout}. The overfitting issue has been a difficult problem in training large
neural network models. The dropout technique has proven to be simple yet very effective
for regularization by preventing complex co-adaptations during training.
In our work,  we define a {\it macro-block} that contains a large number of
  units from the input to a Recurrent Neural Network (RNN).
 Rather than applying dropout to each unit, we apply random dropout
  to each {\it macro-block}.  This algorithm has the effect of applying
  different drop out rates for each layer even if we keep a constant average dropout
  rate, which has better regularization effects.
  In our experiments using Recurrent Neural
  Network-Transducer (RNN-T), this algorithm shows relatively 4.30 \% and
  6.13 \% Word Error Rates (WERs) improvement over the conventional dropout on
   LibriSpeech {\tt test-clean} and {\tt test-other}.
  With an Attention-based Encoder-Decoder (AED) model,  this algorithm shows
  relatively 4.36 \% and 5.85 \% WERs improvement over the conventional
  dropout on the same test sets.
  \vspace{5mm}
\end{abstract}

\section{Introduction}
Deep learning technologies have significantly
improved speech recognition accuracy recently \cite{
c_kim_acssc_2020_00, c_chiu_icassp_2018_00}.  There have been series of remarkable
changes in speech recognition algorithms during the past decade.
These improvements have been obtained by the shift from Gaussian Mixture Model
(GMM) to the Feed-Forward Deep Neural Networks (FF-DNNs), FF-DNNs
to Recurrent Neural Network (RNN) such as the Long Short-Term Memory
(LSTM) networks \cite{S_Hochreiter_neural_computation_1997_00}.
Thanks to these advances, voice assistant devices such as Google Home
\cite{c_kim_interspeech_2017_00}, Amazon Alexa and
Samsung Bixby are widely used at
home environments.
Tremendous amount of research has been conducted in the process of switching
from a conventional speech recognition system consisting of
an Acoustic Model (AM), a Language Model (LM), and a decoder based on
a Weighted Finite State Transducer (WFST) to a complete end-to-end
all-neural speech recognition system
\cite{j_chorowski_nips_2015_00}.

Such notable shifts happened not only in research on model architectures, but also in
research on model robustness as well. In conventional approaches, researchers
have focused on cleaning or transforming speech signals
\cite{e_habets_taslp_2010_00, y_ephraim_tsap_1995_00, c_kim_asru_2009_1,
c_kim_interspeech_2014_00, s_mun_icassp_2019_00, j_chung_odyssey_2020_00} and features \cite{u_h_yapanel_speechcomm_2008,
c_kim_taslp_2016_00}. However, it has been recently observed that data
augmentation \cite{c_kim_icassp_2018_00, c_kim_interspeech_2018_00} is
especially powerful in enhancing the model robustness. Data augmentation during
the training may be helpful in reducing the environmental mismatch between the
training and testing time. The Small Power Boosting (SPB) algorithm
\cite{c_kim_asru_2009_01}  may be considered as an extreme example of reducing
environmental mismatch by intentionally transforming portions of inputs which
are more susceptible to noise or reverberation.

A large number of end-to-end speech recognition
systems are based on the Attention-based Encoder-Decoder (AED)
\cite{j_chorowski_nips_2015_00} and the Recurrent Neural Network-Transducer
(RNN-T) \cite{a_graves_icassp_2013_00} algorithms.
These complete end-to-end systems have started outperforming
conventional WFST-based decoders for large vocabulary speech recognition tasks
\cite{c_chiu_icassp_2018_00}.
Further improvements in these end-to-end speech recognition systems
have been possible thanks to a better choice of target units such as
Byte Pair Encoded (BPE) and {\it unigram language model} \cite{t_kudo_acl_2018_00} subword units,
and an improved training methodologies such as Minimum Word Error Rate (MWER) training
\cite{r_Prabhavalkar_icassp_2018_00}.
In training such all neural network structures, over-fitting has been a major
issue. For improved regularization in training, various approaches have been
proposed \cite{c_kim_icassp_2020_00}. Data-augmentation has been also proved to
be useful in improving model training
\cite{c_kim_interspeech_2017_00, c_kim_icassp_2021_00,
s_park_interspeech_2019_00, s_mun_dcase_2017_00}.
The dropout approach \cite{n_srivastava_jmlr_2014_00} has been applied
to overcome this issue in which both the input and the hidden
units are randomly dropped out for regularization.
This dropout approach has inspired a number of related approaches
\cite{pmlr_l_wan_2013_00, g_larsson_iclr_2017_00, x_gastaldi_iclr_workshop_00}.
In {\it DropBlock} \cite{g_ghiasi_nips_2018_00}, the authors claimed that
dropping out at random is not effective in removing semantic information
when training Convolutional Neural Networks (CNNs) because nearby activations
contain closely related information. Note that in CNN, filters are applied to
nearby elements, thus activation units are spatially correlated.
Motivated by this, they apply a square mask centered around each zero position.
However, since this kind of spatial correlation does not hold for fully connected
feed-forward layers or RNNs, the application of {\it DropBlock} is limited to
CNNs.

In this paper, we present  a new regularization algorithm referred to as  {\it
macro-block dropout}. We define a {\it macro-block} that contains multiple
input units to a neural network layer.  Rather than applying
dropout to each unit, we apply random dropout to each {\it macro-block}.
In our experiments using an RNN-T \cite{a_graves_icassp_2013_00} and an
Attention-based Encoder Decoder (AED) in
Sec. \ref{sec:experimental_results}, this simple algorithm has shown a quite
significant improvement over the conventional dropout approach.
Our contribution in this paper may be summarized as follows. First, by using large
macro-blocks, the ratio of dropped input units in each layer
has large variation even if we keep a constant dropout rate.
This variation in the ratio of dropped units leads to better regularization.
Thus, unlike {\it DropBlock} that relies on spatial correlation in CNNs,
we may apply {\it macro-block dropout} to any kinds
of neural networks such as Feed-Forward (FF) networks and RNNs.
To the best of our knowledge, our work is the first in using big chunks
consisting of large number of input units to RNNs for masking.
Second, we propose a new way of input scaling after applying a mask in {\it macro-block
dropout}. This new scaling approach is related to the fact that the
portion of dropped units significantly varies in {\it macro-block dropout}.
Third, we proposed a very low-cost regularization algorithm.
As will be seen in Sec \ref{sec:experimental_results}, the best performance is
achieved when the number of macro blocks for each layer is only four. This
means that we only need to generate four random values for each layer.
Thus, this {\it macro-block} dropout is very simple with very small computational
requirement.
\section{Related works}
\label{sec:related_works}
%
\begin{figure}
  \centering
  {\includegraphics[width=90mm]{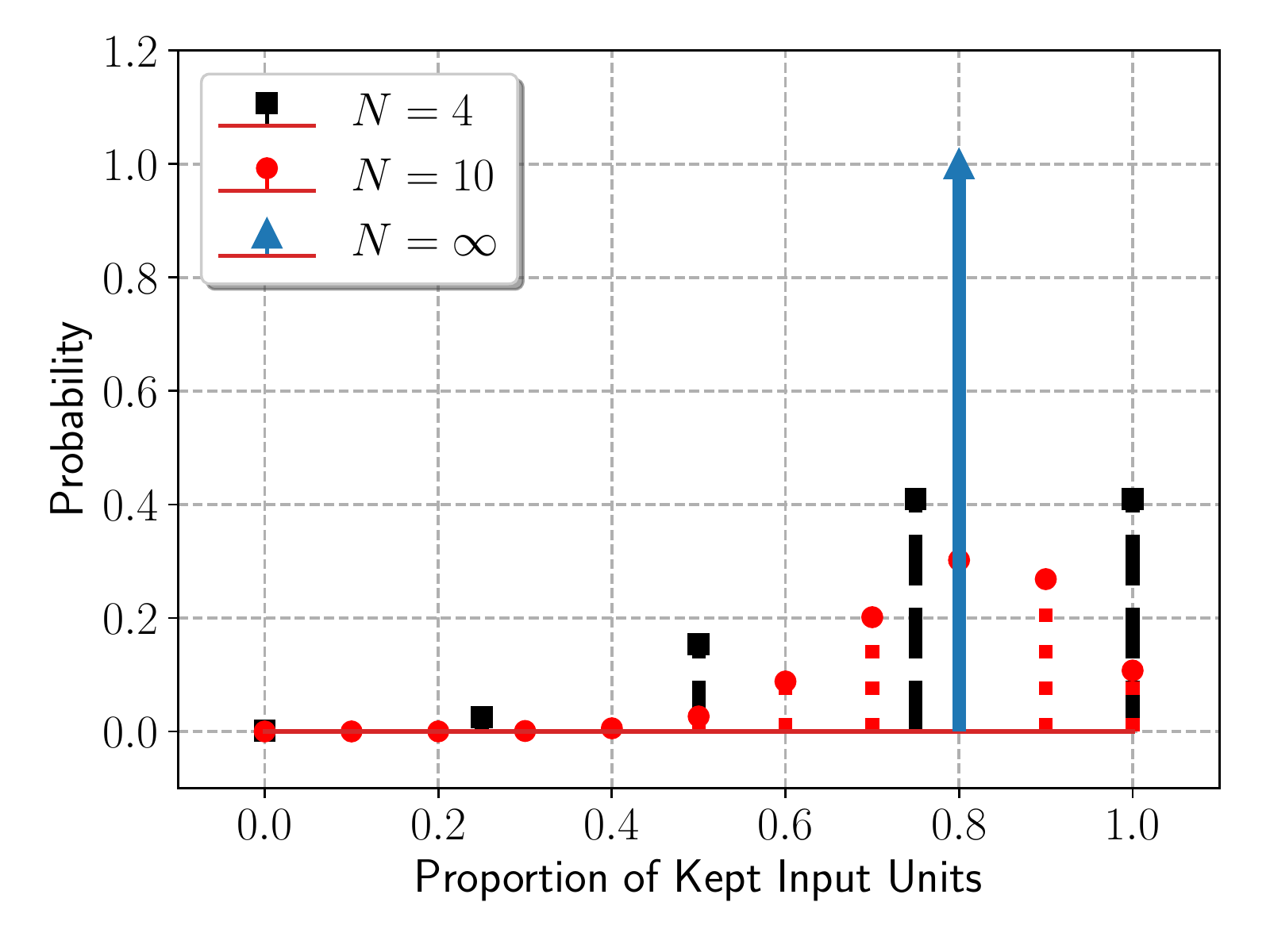}}

  \caption{\label{fig:kept_proportion}
      The Probability Mass Function (PMF) representing the ratio of kept
    input units when the numbers of macro blocks are $N=4$ and $N=10$, respectively.
    In this case, the dropout rate $q$ is assumed to be 0.2.
    When the conventional dropout is applied, the total number of input units to
    an RNN layer is usually from tens of thousands up to several millions.
    In this case, the Probability
    Density Function (PDF) can be approximated by a delta function, which is
    represented by the plot with the legend label of $N=\infty$.
  }
  \vspace{5mm}
\end{figure}
{\it Dropout} is a simple regularization technique to alleviate the overfitting problem by
preventing co-adaptations \cite{n_srivastava_jmlr_2014_00}. When the dimension of
the input to a neural network layer is $\mathbbm{d}_{\mathbf{x}}$, we create
a random mask tensor $\mathbf{m}$ with the same dimension
$\mathbbm{d}_{\mathbf{x}}$. Each element $m \in \mathbf{m}$ follows the
Bernoulli distribution $m  \sim {\text {\it Bernoulli}(1-q)}$, where $q$ is the
dropout rate. Given an input $\mathbf{x}$, the dropout output $\mathbf{x}_{\text{out}}$ is obtained by the following equation:
\begin{align}
  \mathbf{x}_{\text{out}} = \frac{\mathbf{x} \odot \mathbf{m}}{1 - q},
  \label{eq:baseline_dropout}
\end{align}
where $\odot$ is a Hadamard product.
The scaling by $\frac{1}{1-q}$ is applied
to keep the sum of input units the same through this masking process.
In \eqref{eq:baseline_dropout}, we adopt the
{\it ``inverted dropout"} approach rather than the original form of {\it dropout} in
\cite{n_srivastava_jmlr_2014_00} where the scaling is performed during the
inference time.
Before introducing our {\it macro-block dropout}, let us consider the variance
of the dropped ratio of an input layer. When the total number of input
units to a single RNN is $N$, the expected ratio of kept units is given by
$\frac{(1-q)N}{N} = 1-q$, and the standard deviation is given by $\sqrt{\frac{q
(1 - q)}{N}}$. From the central limit theorem, we know that this
distribution can be approximated by a Gaussian distribution.
If we use typical values of $q=0.2$ and $N=10^5$, the standard deviation
becomes $0.00126$, which is very small.  As shown in Fig.
\ref{fig:kept_proportion}, when there are a large number of units, the
distribution is very similar to a delta function centered at $1-q$.
Since the ratio of kept units is always very close to $1-q$, we conclude
that $\frac{\sum \mathbf{x} \odot \mathbf{m}  }{\sum \mathbf{x} } \approx 1-q
$. Thus  we can safely use the fixed scaling of $\frac{1}{1-q}$ to keep the sum
of the input units the same after dropout in \eqref{eq:baseline_dropout}.
Dropout has been turned out to be
especially useful in improving the training of dense network models for image
classification \cite{a_krizhevsky_nips_2012_00}, speech recognition
\cite{g_e_dahl_icassp_2013_00}, and so on.
This dropout approach inspired many other related approaches such as
{\it DropConnect } \cite{pmlr_l_wan_2013_00}, {\it drop-path} \cite{g_larsson_iclr_2017_00},
shake-shake \cite{x_gastaldi_iclr_workshop_00},
{\it ShakeDrop} \cite{y_yamada_ieee_access_2019}, and
{\it DropBlock }\cite{g_ghiasi_nips_2018_00}
regularizations.

%
%
The authors of {\it DropBlock} \cite{g_ghiasi_nips_2018_00} claim that dropping
out inputs to CNNs at random is not effective in removing spatially correlated
information.
In {\it DropBlock}, the {\it zero position}, which is the center of a square mask, is randomly sampled in the same way as the {\it
dropout} in \cite{n_srivastava_jmlr_2014_00}.
It has been reported that {\it DropBlock} significantly outperforms the
baseline {\it dropout} when applied to CNNs.
However, unlike our {\it macro-block dropout}, Bernoulli random numbers are
generated for each unit of inputs to neural networks.  Thus, the ratio
of dropped units has very small variation similar to that of the conventional
dropout case, which is one of the biggest differences from the proposed
{\it Macro-block dropout}.

%
%
\begin{figure}
  \begin{subfigure}{1.0\linewidth}
      \centering
      \resizebox{80mm}{!} {\input{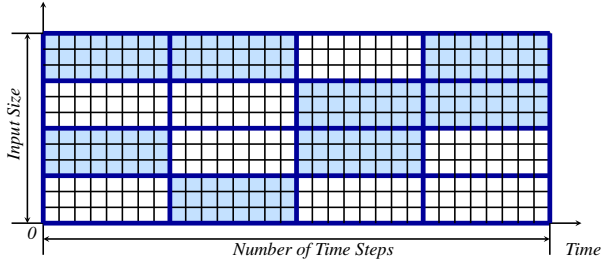}}
    \caption{
      Application of two-dimensional {\it macro-block dropout } \\
      to the input of an RNN.
      \label{fig:2d_macroblock}
    }
  \end{subfigure} \\
  \begin{subfigure}{1.0\linewidth}
      \centering
      \resizebox{80mm}{!} {\input{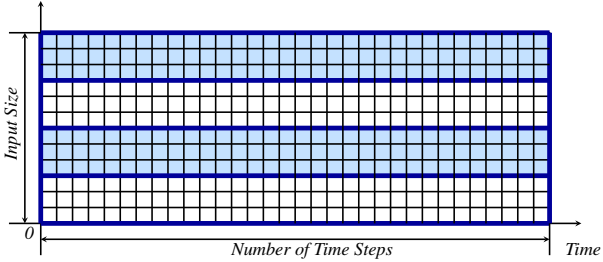}}
    \caption{
      Application of one-dimensional {\it macro-block dropout } \\
      to the input of an RNN.
      \label{fig:1d_macroblock}
    }
  \end{subfigure} \\
  \caption{
Application of {\it macro-block dropout} to the inputs to a Recurrent
    Neural Network (RNN) layer: (\ref{fig:2d_macroblock})  Two-dimensional
    and (\ref{fig:1d_macroblock}) One-dimensional {\it macro-block dropout} cases.
    Each tiny rectangle defined by the grid
    corresponds to each input unit. Larger rectangular chunks are {\it macro-blocks}.
    Region in the light blue color represent {\it macro-blocks} to be dropped out.
  \label{fig:macro_block_to_rnn}}
\end{figure}
\begin{algorithm*}
  \begin{minipage}{1.0\linewidth}
  \caption{Macro-block Dropout}
  \begin{algorithmic}[1]
    \State \textbf{Input}: \emph{Inputs to a layer:} $\mathbf{x}$,
    \emph{the dimension for partitioning:} $\mathbbm{d}_{\text{(par)}}$, \emph{dropout\_rate} $q$, \emph{mode}
    \If {\emph{mode} == \emph{Inference}}
      \State return $\mathbf{x}$
    \EndIf

    \State Creates a random tensor $\mathbf{r}$ with a dimension of
    $\mathbbm{d}_{\text{(par)}}$: \\
      \qquad For each element $r$ of $\mathbf{r}$, $r$
        $\sim$ \emph{Bernoulli}($1 - q$).

    \State Creates a masking tensor $\mathbf{m}$ by resizing $\mathbf{r}$ using
    the nearest-neighbour method to match the dimension of $\mathbf{x}$.

    \State Applies the mask:  \\
      \qquad $\mathbf{x}_m =  \mathbf{x} \odot \mathbf{m}$.

    \State Obtains the output $\mathbf{x}_{\text{out}}$ by scaling $\mathbf{x}_m$ : \\
      \qquad  $\mathbf{x}_{\text{out}} =
          \left|
              \frac{\sum_{\text{all elements}} \mathbf{x} }
                   {\sum_{\text{all elements}} \mathbf{x} \odot \mathbf{m}}
          \right| \mathbf{x}_m $.
  \end{algorithmic}
  \label{algo:macro_block_dropout}
  \end{minipage}
\end{algorithm*}
%
%
%
\section{Macro-block dropout}
%
%
%
%
\subsection{Definition of a macro-block}
\label{sec:def_macro_block}
Let us consider a two-dimensional data $\mathbf{x}$ with the dimension of
$\mathbbm{d}_{\mathbf{x}} = (N_x, N_y)$ that is the input
to a neural-network layer.
{\it Macro-blocks} are constructed by equally partitioning this input
 $\mathbbm{d}_{\mathbf{x}}$ along each axis into a new dimension
of $\mathbbm{d}_{\text{(par)}} = (P_x, P_y)$ with the constraint of
$N_x = P_x M_x$ and $N_y = P_y M_y$ where $M_x$ and $M_y$ are
macro block sizes along the x and y axes.
%
%
%
%

Fig. \ref{fig:macro_block_to_rnn} illustrates examples of {\it macro-blocks}
when this algorithm is applied to  the input of an RNN.
As shown in this figure, the input of an RNN layer is a two-dimensional
variable with the dimension of ($\text{number\_of\_time\_steps}$, $\text{input\_size}$).
Fig. \ref{fig:2d_macroblock} shows the case when this two-dimensional region is
partitioned by $\mathbbm{d}_{\text{(par)}} = (4, 4)$. We refer to this approach
as the two-dimensional {\it macro-block} approach.
As another example, we may also consider the case when
$\mathbbm{d}_{\text{(par)}} = (1, 4)$, which is shown in Fig.
\ref{fig:1d_macroblock}. In this case, the mask pattern does not change
along the time axis. We refer to this approach as the one-dimensional
{\it macro-block} approach.

Unlike the conventional {\it dropout} or the {\it
DropBlock}
approach described in Sec. \ref{sec:related_works}, since there is only a very small
number of macro-blocks in the input, the ratio of kept input units in
a single layer has
very large variation as shown by the stem plot with the legend label of $N=4$
in Fig. \ref{fig:kept_proportion}. For example,
when $\mathbbm{d}_{\text{(par)}} = (1, 4)$ and $q = 0.2$, with the probability
of $0.8^4 = 0.4096$, dropout will not happen at all for a single layer. With the probability of
${4 \choose 2} 0.8^2 0.2 ^ 2 = 0.1536$, half of the input units will be dropped
out. The standard deviation of kept ratio is given by $\sqrt{\frac{q (1 - q)}{4}} = 0.2$. We believe that such large variation is the reason why
{\it macro-block dropout} works for RNNs while {\it DropBlock} is limited to
CNNs.

The Word Error Rates (WERs) using
an RNN-T model with the one-dimensional and the two-dimensional {\it macro-block
dropout} approaches are summarized in Table \ref{tbl:1d_2d_macro_block_result}.
The structure of this  RNN-T model and the experimental configuration are
described in detail in Sec. \ref{sec:speech_recognition_model} and
Sec.  \ref{sec:experimental_results}, respectively. From this result,
we conclude that the one-dimensional {\it macro-block dropout} approach is more
effective than two-dimensional approach for RNNs.
This observation is similar to what is mentioned in \cite{y_gal_nips_2016_00}
for the baseline dropout. The masking using the one-dimensional approach
corresponds to ``the same weight realization for a single input", which is
consistent with the Bayesian interpretation of {\it dropout} for RNNs. This
improvement has been commonly found in other literatures as well \cite{taesup_m_asru_2015_00}.

In obtaining this result,
we choose the partition dimensions $\mathbbm{d}_{\text{(par)}} = (1, 4)$
and $\mathbbm{d}_{\text{(par)}} = (4, 4)$ for one- and two-dimensional cases,
respectively.  We choose these dimensions since the best WERs for
one- and two- dimensional {\it macro-block dropout} cases are obtained with
these dimensions in our experiments on the {\it LibriSpeech} corpus.
\begin{table}[!htbp]
  \renewcommand{\arraystretch}{1.3}
  \centering
        \caption{\label{tbl:1d_2d_macro_block_result}
        Word Error Rates (WERs) with the RNN-T model described in
        Sec. \ref{sec:speech_recognition_model}
        using the one-dimensional {\it macro-block dropout} of
        $\mathbbm{d}_{\text{(par)}} = (1, 4)$, and the two-dimensional
        {\it macro-block dropout} of $\mathbbm{d}_{\text{(par)}} = (4, 4)$.
        In these experiments, the dropout rate of 0.2
        is used since the best WER in each case is obtained at this rate.
        }
  \begin{minipage}{0.9\linewidth}

\smallskip
    \begin{tabularx}{\textwidth}{ c * 4 {Y}}
\toprule
      \multirow{2}{*}{Test Set} &
      \multirow{2}{*}{\specialcell{Baseline \\ Dropout}} &
      \multicolumn{2}{c}{Macro-Block Dropout} \\
      \cmidrule{3-4}
      & & \specialcell{1-D : $(1, 4)$}
      &   \specialcell{2-D : $(4, 4)$} \\
\midrule
      {\tt  test-clean} &  {\bf 3.95 } \% &  \textcolor{blue}{ \bf 3.78} \%
      &  3.92 \%     \\
      {\tt test-other} &  {\bf 12.23} \% &  \textcolor{blue}{\bf 11.48} \% &  11.50  \%   \\
\bottomrule
\end{tabularx}
  \end{minipage}
\end{table}
\subsection{Application of dropout to macro-blocks}
\label{sec:macro_block_dropout_algorithm}
Having defined the required terms in Sec. \ref{sec:def_macro_block}, we
proceed to explain the algorithm in detail in this section.
The entire algorithm is summarized in Algorithm \ref{algo:macro_block_dropout}.
During the inference time, {\it macro-block dropout} is not applied as the
original dropout.
During the training time, we create a random tensor $\mathbf{r}$ whose
dimension is
$\mathbbm{d}_{(\text{par})}$. This tensor is created from the {\it Bernoulli}
distribution with the probability of one given by $1 - q$,
where $q$ is the dropout probability.
This $\mathbf{r}$ is then resized to match the dimension of the input $\mathbf{x}$. For
simplicity, this resize operation is performed using the
well-known {\it nearest-neighborhood} interpolation approach.

The scaling factor $r$ is given by the following equation:
\begin{align}
  r = \left|\frac{\sum_{\text{all elements}} \mathbf{x} }
                 {\sum_{\text{all elements}} \mathbf{x} \odot \mathbf{m}
                 } \right|.
    \label{eq:scaling_factor}
\end{align}
We apply the absolute value operation in \eqref{eq:scaling_factor}, because
the sign of the numerator and the denominator of \eqref{eq:scaling_factor} may
be different when $\mathbf{x}$ is the output  of an RNN such as an LSTM or a GRU.
More specifically, the hidden output of an LSTM is given by the following
equation \cite{S_Hochreiter_neural_computation_1997_00, c_kim_acssc_2020_00}:
\begin{align}
  \mathbf{h}_{[m]} = \mathbf{o}_{[m]} \odot \sigma_h(\mathbf{c}_{[m]}),
  \label{eq:hidden_output}
\end{align}
where $m$ is a time index, $\odot$ is the Hadamard product, $\sigma_h(\cdot)$ is
the hyperbolic tangent function, $\mathbf{o}_{[m]}$ is the output-gate value,
and $\mathbf{c}_{[m]}$ is the cell value, respectively. From \eqref{eq:hidden_output}, it is obvious that $\mathbf{h}_{[m]}$
may have both positive and negative values, since the range of of $\sigma_h$ is
between -1 and 1. In our speech recognition experiments,
it is observed that performance is slightly worse if this absolute value
operation is not applied in \eqref{eq:scaling_factor}. In performing division
in \eqref{eq:scaling_factor}, we employ {\it ``a safe division"} implemented
by the {\tt tf.math.divide\_no\_nan} {\tt Tensorflow} \cite{m_abadi_usenix_2016} API to prevent division by zero.

We observe that the scaling in  \eqref{eq:scaling_factor} is more effective
than the simple scaling of $\frac{1}{1-q}$ used in the baseline
dropout in \eqref{eq:baseline_dropout}
since $\frac{\sum \mathbf{x} \odot \mathbf{m}  }{\sum \mathbf{x} } \approx 1-q$
does not hold with {\it macro-block dropout} because of the large variance
in the ratio of kept input units.
Table \ref{tbl:scaling_macro_block_result} summarizes WERs obtained with the
conventional scaling of $\frac{1}{1-q}$ and the scaling in
\eqref{eq:scaling_factor} on the {\it LibriSpeech} {\tt test-clean} and {\tt
test-other} sets. We use the RNN-T model that will be described in
Sec. \ref{sec:experimental_results}. For {\it macro-block dropout}, we employ the
one-dimensional approach with the partition dimension of
$\mathbbm{d}_{\text{(par)}} = (1, 4)$. The experimental configuration in obtaining
these results will be described in Sec. \ref{sec:experimental_results}.

\begin{table}[!htbp]
  \renewcommand{\arraystretch}{1.3}
  \centering
        \caption{\label{tbl:scaling_macro_block_result}
        Word Error Rates (WERs) with the RNN-T model described in Sec.
        \ref{sec:speech_recognition_model}
        using the scaling suggested by \eqref{eq:scaling_factor} and
        $\frac{1}{1-q}$. The dropout rate is 0.2 and the partition dimension for
        the 1-dimensional {\it macro-block dropout} is
        $\mathbbm{d}_{\text{(par)}} = (1, 4)$.
        }
  \begin{minipage}{0.9\linewidth}
\smallskip
    \begin{tabularx}{\textwidth}{ c * 4 {Y}}
\toprule
      \multirow{2}{*}{Test Set} &
      \multirow{2}{*}{\specialcell{Baseline \\ Dropout}} &
      \multicolumn{2}{c}{\specialcell{1-D Macro-Block \\ Dropout: $(1, 4)$}} \\
      \cmidrule{3-4}
      &  &  \specialcell{Scaling \\ using \eqref{eq:scaling_factor}}
      &  \specialcell{Scaling \\ using $\frac{1}{1-q}$} \\
\midrule
      {\tt  test-clean} &  {\bf 3.95 } \% &  \textcolor{blue}{ \bf 3.78} \%
      &  4.04 \%     \\
      {\tt test-other} &   {\bf 12.23} \% &  \textcolor{blue}{\bf 11.48} \%
      &  11.50  \%   \\
\bottomrule
\end{tabularx}
  \end{minipage}
\end{table}
\begin{table*}[!htbp]
  \renewcommand{\arraystretch}{1.3}
  \centering
        \caption{\label{tbl:macro_block_result}
        Word Error Rates (WERs) with the RNN-T model and the Attention-based
        Encoder-Decoder (AED) model described in Sec.
        \ref{sec:speech_recognition_model}
        using the baseline dropout and the
        one-dimensional {\it maro-block
        dropout} approaches.  In these experiments, the dropout rate of 0.2
        is used since the best WERs for both approaches are obtained at this rate.
        }
  \begin{minipage}{0.9\linewidth}

\smallskip
    \begin{tabularx}{\textwidth}{ c * 8 {Y}}
\toprule
    \multirow{3}{*}{Model} &
    \multirow{3}{*}{Test Set} &
      \multirow{3}{*}{\specialcell{Baseline \\ Dropout}}  &
      \multicolumn{4}{c}{\specialcell{One-Dimensional Macro-Block Dropout \\
      with different $\mathbbm{d}_{\text{(par)}}$  }}  \\
      \cmidrule{4-7}
      &        &        & $(1, 3)$  &  $(1, 4)$  & $(1, 5)$  &  $(1, 10)$  \\ 
\midrule
    \multirow{2}{*}{RNN-T} &
      {\tt  test-clean} &  {\bf 3.95 } \% &   4.11 \%
      & \textcolor{blue}{ \bf 3.78} \% &  3.88 \% & 3.94  \%   \\
      &{\tt test-other} &   {\bf 12.23 } \% &  11.57 \%
      & \textcolor{blue}{\bf 11.48} \% &  11.52  \% & 11.50 \%  \\
\hline
      \multirow{2}{*}{\specialcell{Attention-based\\Encoder Decoder}} &
      {\tt  test-clean} &  {\bf 3.67 } \% &   3.66 \%
      & \textcolor{blue}{ \bf 3.51} \% &  3.54 \% & 3.61  \%   \\
      &  {\tt test-other} &   {\bf 11.62 } \% &  11.20 \%
      & \textcolor{blue}{\bf 10.94} \% &  10.98  \% & 11.07 \%  \\
\bottomrule
\end{tabularx}
  \end{minipage}
\end{table*}

\section{Speech recognition model}
\label{sec:speech_recognition_model}
%
%
For speech recognition experiments, we employed an RNN-T speech recognizer
and an Attention-based Encoder Decoder (AED).
Our speech recognition system is built {\it in-house} using
{\tt Keras} models \cite{f_chollet_keras_2015_00} implemented for
{\tt Tensorflow} 2.3 \cite{m_abadi_usenix_2016}.
The RNN-T structures have three major components: an encoder (also known as
a transcription network), a prediction network, and a joint network.
In our implementation, the encoder consists of six layers of bi-directional
LSTMs interleaved with 2:1 max-pooling layers \cite{m_ranzato_cvpr_2007_00}
in the bottom three layers.  Thus, the overall temporal sub-sampling factor is 8:1
because of these three 2:1 max-pooling layers. The prediction network consists
of two layers of uni-directional LSTMs.  The unit size of all these LSTM layers is
1024. {\it Macro-block dropout} is applied
to all the inputs of each LSTM layer of the encoder and the prediction network
except the first layer of the encoder.
From the transducer output, a linear embedding vector with a dimension of $621$
is obtained and fed back into the prediction network. The AED
model has three components: an
encoder, a decoder, and an attention block. In our implementation, the encoder
of the AED model is identical to the encoder structure of the RNN-T model
explained above. As a decoder, we use a single layer of uni-directional LSTM whose unit size is 1024.

The loss employed for training the RNN-T model is a combination of the
Connectionist Temporal Classification (CTC) loss \cite{a_graves_icml_2006_00}
applied to the {\it encoder} output and the RNN-T loss
\cite{a_graves_icassp_2013_00} applied to the full network,
 which is represented by the following:
\begin{align}
  \mathbbm{L}_{\text{CTC-RNN-T}}  =  \mathbbm{L}_{\text{CTC}} + \mathbbm{L}_{\text{RNN-T}}.
    \label{eq:joint_ctc_rnn_t_loss}
\end{align}
We refer to this loss in \eqref{eq:joint_ctc_rnn_t_loss} as the {\it joint CTC-RNN-T loss}.
For the AED model, we employ the joint {\it CTC-CE loss}, which is given by:
\begin{align}
  \mathbbm{L}_{\text{CTC-CE}}  =  \mathbbm{L}_{\text{CTC}}
  + \mathbbm{L}_{\text{CE}},
    \label{eq:joint_ctc_ce_loss}
\end{align}
as in \cite{c_kim_acssc_2020_00}.
 For better stability during the training, we use the gradient clipping by
global norm \cite{r_pascanu_icml_2013}, which is implemented in {\tt
Tensorflow} as the {\tt tf.clip\_by\_global\_norm } API.
To obtain further improvement in speech recognition accuracy, we incorporate
an improved shallow-fusion technique.  In this approach, we subtract log prior
probabilities of each label
obtained from the transcript of the speech recognition training database. This
idea is initially described in \cite{n_kanda_taslp_2017_00}. Our formulation is based on
our earlier work in \cite{c_kim_interspeech_2019_00}:
\begin{align}
  \log p_{\text{sf}} & \left( y_l \big|  \ssf{x}_{[0:m]}  \right)  =
    \log p \left(y_l \big| \ssf{x}_{[0:m]}, \hat{y}_{0:l} \right) \nonumber \\
  &  - \lambda_{p} \log p_{\text{prior}} \left(  y_l \right)
    + \lambda_{\text{lm}}  \log p_{\text{lm}} \left( y_l \big|
    \hat{y}_{0:l}\right),
    \label{eq:improved_shallow_fusion}
\end{align}
where $y_l$ is the output at the output label index
$l$, $\ssf{x}_{[0:m]}$ is the input feature vector sequence from the zero-th frame
up to the (m-1)-st frame, and $\hat{y}_{0:l}$ is the estimated output label
sequence from the output index zero up to $l-1$. $\lambda_p$ and
$\lambda_{\text{lm}}$ are constant weighting coefficients for the log prior
probability denoted by $\log p_{\text{prior}} \left(  y_l \right)$ and
the log probability from the Language Model (LM) denoted by $\log p_{\text{lm}}
\left( y_l \big|\hat{y}_{0:l}\right) $, respectively. $\log p \left(y_l \big|
\ssf{x}_{[0:m]}, \hat{y}_{0:l} \right)$ is the log probability from the speech
recognition model.
In our experiments, we use $\lambda_p$ of 0.002 and $\lambda_{\text{lm}}$ of
0.48 respectively as in \cite{c_kim_interspeech_2019_00}.

%
%
\section{Experimental results}
\label{sec:experimental_results}
\begin{table}[!htbp]
  \renewcommand{\arraystretch}{1.3}
  \centering
        \caption{\label{tbl:shallow_fusion}
        Word Error Rates (WERs) with the Attention-based Encoder Decoder (AED) model
        described in Sec. \ref{sec:speech_recognition_model}
        with the improved shallow fusion in \eqref{eq:improved_shallow_fusion} with
        a Transformer LM \cite{a_vaswani_nips_2017_00}.
        }
  \begin{minipage}{0.9\linewidth}
\smallskip
    \begin{tabularx}{\textwidth}{ c * 3 {Y}}
\toprule
      {Test Set} &
      \specialcell{Baseline Dropout} &
      \specialcell{Macro-Block Dropout}  \\
\midrule
      {\tt  test-clean} &  {\bf 2.44 } \% &  \textcolor{blue}{ \bf 2.37} \% \\
      {\tt test-other}  &  {\bf 7.87 } \% &  \textcolor{blue}{\bf 7.42} \% \\
\bottomrule
\end{tabularx}
  \end{minipage}
\end{table}
In this section, we explain experimental results using the {\it macro-block
dropout} approach with the RNN-T and the AED model described in Sec.
\ref{sec:speech_recognition_model}.
For training, we used the entire 960 hours LibriSpeech \cite{v_panayotov_icassp2015} training
set consisting of 281,241 utterances. For evaluation, we used the
official 5.4 hours {\tt test-clean} and 5.1 hours {\tt test-other} sets
consisting of 2,620 and 2,939 utterances respectively.
The pre-training stage has some similarities to our previous work
in \cite{c_kim_asru_2019_01}. In this pre-training stage, the number of
LSTM layers in the encoder increased at every 10,000-steps starting from two
LSTM layers up to six layers. We use an Adam
optimizer \cite{d_p_kingma_iclr_2015_00} with the initial learning rate of
0.0003, which is maintained for the entire pre-training state and one full epoch
  after finishing the pre-training stage.
After this step, this learning rate decreases exponentially with a decay rate
of 0.5 for each epoch.
$\mathbf{x}[m]$ and $\mathbf{y}_l$ are the input {\it power-mel
filterbank} feature of size 40 and the output label, respectively.
$m$ is the input frame index and $l$ is the decoder output
step index.
We use the {\it power-mel filterbank} feature instead of the more
commonly used {\it log-mel filterbank} feature based on our previous
experimental results
\cite{c_kim_interspeech_2019_00,  c_kim_asru_2019_00, c_kim_asru_2019_01}. In
out {\it power-mel filterbank} feature, we employ the power-law nonlinearity
with the power coefficient of $\frac{1}{15}$, which is suggested in
\cite{c_kim_taslp_2016_00, c_kim_phd_thesis_2010}.
For better regularization in training, we apply the {\it SpecAugment} as
a data-augmentation technique in all the experiments
\cite{s_park_interspeech_2019_00}.
In our experiments with different dropout rates ranging from 0.0 to 0.5,
for both the conventional
{\it dropout}  and the {\it macro-block dropout}
approaches, the best WERs are obtained when the dropout rate $q$ is 0.2.

Table \ref{tbl:macro_block_result} summarizes the
experimental results with conventional dropout and {\it macro-block}
dropout approaches using the RNN-T and AED models. In the case of the {\it macro-block} dropout approach, we conducted
experiments with four different partition sizes $\left(\mathbbm{d}_{\text{(par)}}
= \{(1, 3), (1, 4), (1, 5), (1,10)\}\right)$ with the one-dimensional
masking pattern that is shown in Fig. \ref{fig:1d_macroblock}. From this table,
we observe that the best WER is obtained when the number of blocks is four
($\mathbbm{d}_{\text{(par)}} = (1, 4)$).
As shown in this table, when the RNN-T model is employed, the {\it macro-block
dropout} algorithm shows relatively 4.30 \% and 6.13 \% WER
improvements over the conventional
  dropout approach on the {\it LibriSpeech} {\tt test-clean} and {\tt test-other}
  sets, respectively.
In the same table, we observe that the performance improvement using the AED model
is also similar. We obtain 4.36 \% and 5.85 \% relative
WER improvements on the same sets, respectively. 
Finally, we apply the improved shallow fusion in \ref{eq:improved_shallow_fusion}
to further improve the performance. Table \ref{tbl:shallow_fusion} shows WERs
obtained with the AED model using the improved shallow fusion in
\eqref{eq:improved_shallow_fusion} with a Transformer LM
\cite{a_vaswani_nips_2017_00}. As shown in this Table,
{\it macro-block dropout} shows 2.86 \% and 5.72 \% relative WER improvement on
the {\it LibriSpeech} {\tt test-clean} and {\tt test-other} sets.
%
%
%
%
%
%
%
%
%

\section{Conclusions}

In this paper, we describe a new regularization algorithm referred to as {\it macro-block
dropout}. In this approach, rather than applying dropout to each input unit to
RNN layers, we apply a random mask to a bigger chunk referred to as a {\it macro-block}. We
propose an improved way of performing scaling for better performance with {\it
macro-block dropout} in Sec. \ref{sec:macro_block_dropout_algorithm}. In
experiments using RNN-T and AED models, we obtain significantly better
results with {\it macro-block dropout} compared to the conventional dropout approach.
We compare the variance
of the ratio of input units that are not dropped with the conventional dropout approach and with {\it macro-block
dropout} approach. We observe that this variance is significantly larger with {\it
macro-block dropout}. We believe this larger variance helps
regularization during training.
%

\vspace{5mm}
{
\ninept
\bibliographystyle{IEEEtran}
\bibliography{common_bib_file}
}

\end{document}